\documentclass[sigconf]{acmart}

\acmConference[]{}{}{}

\AtBeginDocument{%
  }

\usepackage{makecell}
\usepackage{listings}
\begin{document}

\title{Winning Amazon KDD Cup'24}


\author{Chris Deotte}
\authornote{All authors contributed equally to this research.}
\email{cdeotte@nvidia.com}
\affiliation{%
  \institution{NVIDIA}
  \country{USA}
}

\author{Ivan Sorokin}
\authornotemark[1]
\email{isorokin@nvidia.com}
\affiliation{%
  \institution{NVIDIA}
   \country{Finland}
}

\author{Ahmet Erdem}
\authornotemark[1]
\email{aerdem@nvidia.com}
\affiliation{%
  \institution{NVIDIA}
  \country{Türkiye}
}

\author{Benedikt Schifferer}
\authornotemark[1]
\email{bschifferer@nvidia.com}
\affiliation{%
  \institution{NVIDIA}
  \country{Germany}
}

\author{Gilberto Titericz Jr}
\authornotemark[1]
\email{gtitericz@nvidia.com}
\affiliation{%
  \institution{NVIDIA}
  \country{Brazil}
}

\author{Simon Jegou}
\authornotemark[1]
\email{sjegou@nvidia.com}
\affiliation{%
  \institution{NVIDIA}
  \country{France}
}

\renewcommand{\shortauthors}{Deotte et al.}

\begin{abstract}
    This paper describes the winning solution of all 5 tasks for the Amazon KDD Cup 2024 \textit{Multi Task Online
Shopping Challenge for LLMs}. The challenge was to build a useful assistant, answering questions in the domain of online shopping. The competition contained 57 diverse tasks, covering 5 different task types (e.g. multiple choice) and across 4 different tracks (e.g. multi-lingual).

Our solution is a single model per track. We fine-tune Qwen2-72B-Instruct on our own training dataset. As the competition released only 96 example questions, we developed our own training dataset by processing multiple public datasets or using Large Language Models for data augmentation and synthetic data generation. We apply \textit{wise-ft} to account for distribution shifts and ensemble multiple LoRA adapters in one model. We employed \textit{Logits Processors} to constrain the model output on relevant tokens for the tasks. \textit{AWQ 4-bit Quantization} and \textit{vLLM} are used during inference to predict the test dataset in the time constraints of 20 to 140 minutes depending on the track.

Our solution achieved the first place in each individual track and is the first place overall of Amazon's KDD Cup 2024.

\end{abstract}

\ccsdesc[500]{Computing methodologies~Natural language generation}
\ccsdesc[300]{Computing methodologies~Machine translation}
\ccsdesc[300]{Computing methodologies~Information extraction}

\keywords{Large Language Models, LLM, Shopping Assistant, KDD Cup, Multi Task Learning, Multi-Lingual}


\maketitle

\section{Introduction}
The capabilities of Large Language Models (LLMs) have significantly improved in the last years and they have become popular due to their easiness to use. Users can interact with the systems in natural language. The LLMs excel on a variety of tasks, such as general reasoning, math questions, coding, etc. Many systems are getting updated by adding a LLMs to make them easier to use and/or providing more functionality. (Online) shopping is a large domain with billions of users and high economic output. The Amazon KDD Cup 2024 \cite{kddcup2024} is designed to evaluate LLMs to be a useful shopping assistant.

 \begin{figure}[h!]
    \centering
    \includegraphics[width=0.45\textwidth]{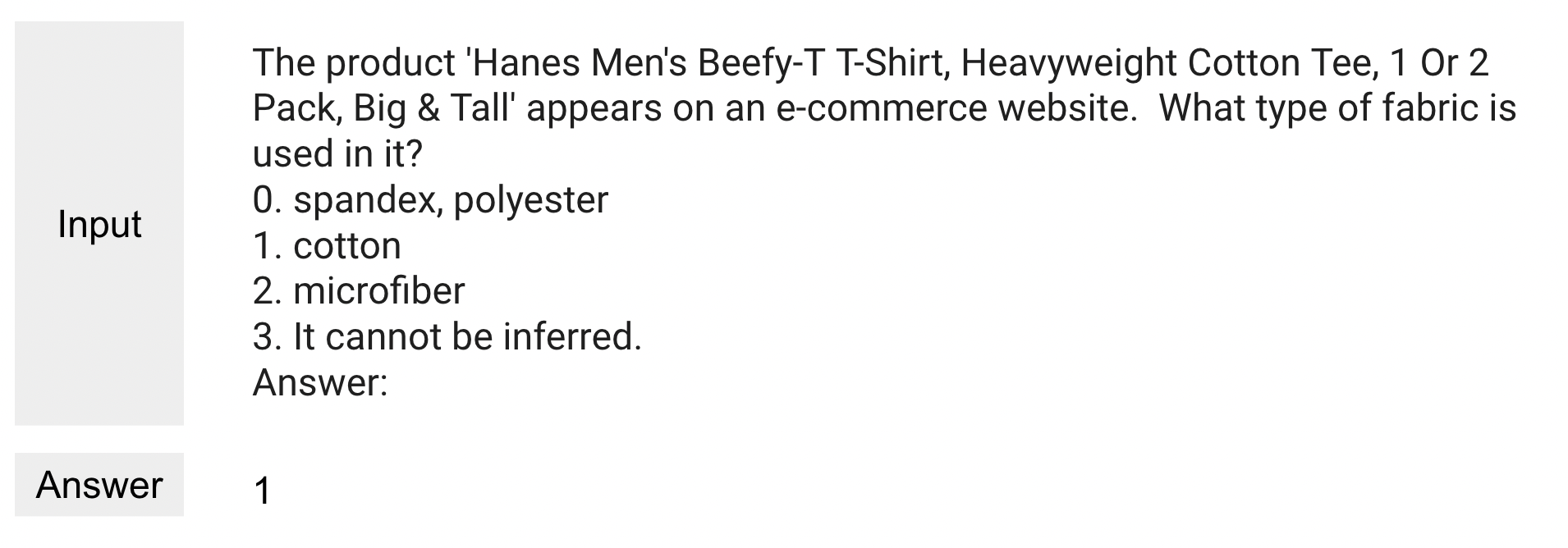}
    \caption{One example of the development dataset. It is a multiple choice question answering tasks for understanding shopping concepts.}
    \label{fig:example}
\end{figure}

Amazon developed an evaluation dataset \textit{ShopBench}, containing approx. 20,000 questions across 57 different tasks covering 5 task types (e.g. retrieval), to test LLMs capabilites in the online shopping domain (see an example in Figure \ref{fig:example}). The competition had 5 different tracks, which evaluates different aspects such as shopping knowledge understanding or user behavior alignment. The 5th track was the overall track containing all 20,000 questions. The competition was organized as a code competition in which participants have no access to the ShopBench dataset and instead they have to submit their model.

 \begin{figure}[h!]
    \centering
    \includegraphics[width=0.45\textwidth]{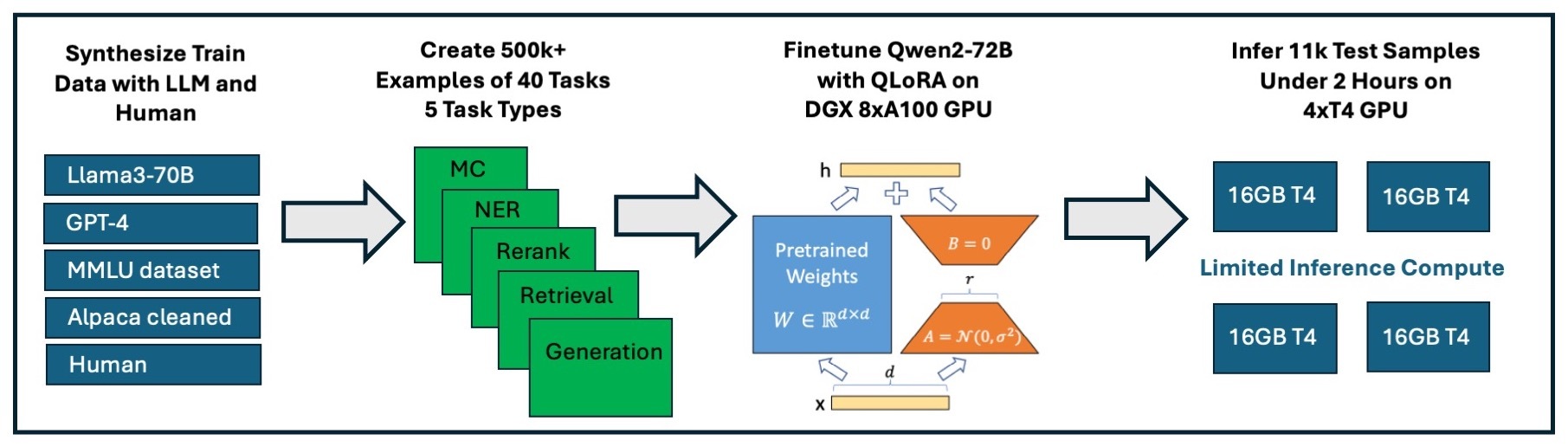}
    \caption{High-Level Oerview of our pipeline for the KDD Cup 2024}
    \label{fig:solution_overview}
\end{figure}

Our team from NVIDIA won all 5 tracks (see Table \ref{tab:leaderboard}). This paper describes our final solution and an ablation study on our experiments.
Our solution is based on a single model per track, which shares following techniques:
\begin{enumerate}
 \item \textbf{Developing a Training Dataset}: As the hosts did not provide a training dataset, we processed many multiple public datasets and enriched it by prompting Large Language Models to generate a training dataset
 \item \textbf{Fine-Tuning a Base Model}: We fine-tuned Qwen2-72B-Instruct \cite{yang2024qwen2technicalreport}
 \item \textbf{Additional Optimization}: We applied multiple techniques to maximize our solution by prompt engineering, ensembling multiple adapters, addressing distribution shift with \textit{wise-ft} and constraining the model with a \textit{Logits Processors}
 \item \textbf{Optimizing Inference}: As the competition had compute and time constraints, we optimized our inference code with 4-bit quantization and vLLM.
\end{enumerate}

We describe our methods in details in the following sections \ref{sec:datasets} and \ref{sec:model}. Section \ref{sec:results} will compare our solutions and share more experiments we ran during the competition as a small ablation study.

\section{Amazon KDDCup 2024: Multi Task Online Shopping Challenge for LLMs}

Amazon hosted the KDD Cup 2024 for \textit{Multi Task Online Shopping Challenge for LLMs} \cite{kddcup2024}. They developed a test dataset, called \textit{ShopBench}, containing ~20,000 questions across 57 tasks. A development dataset of 96 question of only 18 different tasks were shared with the participants for the competition. 
The KDD Cup 2024 is designed as a code competition. Participants submit model weights with code which will be evaluated on infrastructure provided by Amazon. Participants have no access to the test dataset and can build solutions based on the 96 development questions. They receive only the scores on the full \textit{ShopBench} dataset via the leaderboard.
A submission is evaluated on 4x NVIDIA T4 GPUs with each 16 GB GPU memory within a runtime limit (see Table \ref{tab:runtime_limit}).

\begin{table}[]
\caption{Runtime limit in minutes per track in phase 2}
\begin{tabular}{l|lllll}
Phase   & Track 1 & Track 2 & Track 3 & Track 4 & Track 5 \\ \hline
Phase 2 & 70      & 20      & 30      & 20      & 140    
\end{tabular}
\label{tab:runtime_limit}
\end{table}

Participants have to address following challenges:
\begin{enumerate}
 \item \textbf{No training dataset}: Participants have access to only 96  examples.
 \item \textbf{Hidden tasks}: The development dataset contains only 18 out of 57 tasks. Therefore, the solution has to generalize to the unknown tasks.
 \item \textbf{Time and compute constrains}: Solutions have to run within a runtime limit on 4x NVIDIA T4 GPUs with each 16 GB memory (see Table \ref{tab:runtime_limit})
\end{enumerate}

The competition contains 5 tracks:
\begin{itemize}
    \item \textbf{Shopping Concept Understanding}: Understanding shopping concepts (e.g. brands, product lines, attributes, etc.)
    \item \textbf{Shopping Knowledge Reasoning}: Reasoning ability about products or product attributes (e.g. total amount in a product pack, are two products compliments or substitutes, etc.)
    \item \textbf{User Behavior Alignment}: Understanding user behavior in online shopping (e.g. implicit information by user click stream
    \item \textbf{Multi-lingual Abilities}: Shopping concept understanding and user behavior alignment across different languages
    \item \textbf{Overall}: A final track which combines all 4 tracks
\end{itemize}

The evaluation dataset is based on 5 task types:
\begin{itemize}
    \item \textbf{Multiple Choice}: Only one correct answer. Evaluation metric is accuracy.
    \item \textbf{Ranking}: Input contains multiple candidates and the model should provide an ordered list. Evaluation metric is nDCG.
    \item \textbf{Named Entity Recognition (NER)}: Extract pieces of text given an entity type. Evaluation metric is Micro-F1. 
    \item \textbf{Retrieval}: Select candidates from a list which satisfy the requirements. Evaluation metric is Hit@3
    \item \textbf{Generation}: There are a diverse set of generation tasks depending on the task (e.g. translation). Evaluation metrics are ROUGE-L, BLEU or cosine similarity of sentence embedding.
\end{itemize}

A track can contain one or multiple task types. The final score of a track is calculated by averaging across all questions because each evaluation metric is between 0-1. The overall challenge score is determined by the sum of position per track.  

For every question, the requirement is to generate text, which is parsed by Amazon's evaluation script. The solution has to follow the prompt instructions (e.g. \textit{return 3 candidates IDs separated by a comma}). If the evaluation script is not able to parse the generated text, then the score will be 0 for this question.

The competition was organized in 2 phases. The organizer shared that Phase 2 contains harder samples and tasks than Phase 1. They increased the compute resources from 2x NVIDIA T4s to 4x NVIDIA T4s for phase 2.

\section{Training Dataset}
\label{sec:datasets}

Amazon shared multiple eCommerce datasets with participants, which are related to the ShopBench dataset, but do not have the same structure. We created our training dataset by processing multiple datasets to have a similar structure as the 18 tasks from ShopBench development dataset. In addition, we developed new tasks. Finally, we augmented the dataset by prompting LLaMa3-70B-Instruct \cite{llama3modelcard} and GPT-4 \cite{openai2024gpt4technicalreport} for more diversity or infer missing information (e.g. product type, category). A detailed overview can be found in Appendix \ref{sec:training_datasets}.

\subsection{Real Datasets}

We utilized multiple data sources, including non e-commerce datasets such as MMLU and Alpaca-Cleaned. Samples from these datasets were transformed into the instruction prompts.

\begin{description}

\item[Amazon-M2] \cite{jin2023amazonm2multilingualmultilocaleshopping} - A multi-lingual Amazon session dataset with rich meta-data used for KDD Cup 2023.

\item[Amazon Reviews 2023] \cite{hou2024bridging} - A large scale Amazon Review Dataset with rich features and over 500M reviews across 33 categories.

\item[NingLab/ECInstruct] \cite{peng2024ecellm} - instruction dataset covers 116,528 samples from 10 real and widely performed e-commerce tasks of 4 categories.

\item[ESCI-data] \cite{reddy2022shopping} - Shopping Queries dataset provides a list of up to 40 potentially relevant results, together with ESCI relevance judgements (Exact, Substitute, Complement, Irrelevant) indicating the relevance of the product to the query.

\item[MMLU] \cite{hendryckstest2021} \cite{hendrycks2021ethics} - massive multitask test consisting of 16k multiple-choice questions and auxiliary 100k multiple-choice training questions from ARC, MC\_TEST, OBQA, RACE, etc.

\item[Alpaca-Cleaned] \cite{alpaca} -  a cleaned version of the original Alpaca Dataset released by Stanford.

\end{description}

\subsection{Synthetic Datasets}

To further improve diversity of dataset we utilized the synthetic data generation (SDG) pipelines. In general, we used three different methods.

We prompt LLM to construct the tasks specific prompts from the seed data. For example, we rephrase the original tasks from \textit{NingLab ECInstruct} dataset. These tasks include various information about the product (title, description, attributes) and we combine all of them into one prompt. (see Table \ref{tab:training_dataset} Dataset No 11-19).

Before constructing a task specific prompt, we extract the correct labels from the seed data using LLM. For example, we extract the product type, categories or attributes first and then construct the question. (see Table \ref{tab:training_dataset} Dataset No 1,20).

We used GPT-4 to generate the instructions with different wordings, and then used it to construct MC tasks from \textit{ESCI-data} dataset. The correct answer was randomly selected from the E entries, the remaining options were selected from the entries with S/C/I labels (see Table \ref{tab:training_dataset} Dataset No 21-26).

\section{Model}
\label{sec:model} 

\subsection{Prompt Template}
We explored both zero shot LLM models and fine-tuned LLM models. Our final winning solution achieving our best model accuracy is fine-tuned. See Table \ref{tab:foundation_models} for a comparison.

When using zero shot with an instruction tuned LLM, we found it helpful to use both the system role and user role when formatting prompts. Designing better prompts improved the zero shot model's performance.

When fine-tuning, we found that the prompt was not as important because the model is fine-tuned to exhibit a certain behavior given whatever prompt we choose to train with. 

One technique of our fine-tuned models used is to include an instruction to the model identifying which of the 5 task types the model is solving. Then during inference, we used a heuristic rule classifier which determined question task type and included this is the system role's instruction prompt. Specifically, we used the following template.

\begin{lstlisting}[language=python, caption=System prompt template with task type,label={lst:sprompt}]
system_prompt = "You are a helpful online shopping assistant. Your task is {task_type}."
\end{lstlisting}

\subsection{Fine-Tuning Qwen2}
We fine-tuned \textit{Qwen/Qwen2-72B-Instruct} \cite{yang2024qwen2technicalreport} on our developed training dataset using 8x NVIDIA A100 with each 80GB GPU memory. Training on 500k examples takes around 24 hours. We used the library axolotl \footnote{https://github.com/axolotl-ai-cloud/axolotl} and bitsandbytes \footnote{https://huggingface.co/docs/transformers/main/en/quantization/bitsandbytes} with QLoRA \cite{dettmers2023qloraefficientfinetuningquantized} with 4-bit quantization and bfloat16 \cite{8877390}. The library applies \textit{Multipack (Sample Packing)} \footnote{https://github.com/axolotl-ai-cloud/axolotl/blob/main/docs/multipack.qmd}, concatenating multiple sequences into one batch to increase training throughput. Table \ref{tab:hyperparameters} provides an overview of the hyperparameters.

\begin{table}[]
\caption{Model hyperparameters}
\begin{tabular}{lr}
\bf{Hyperparameter}              & \bf{Value}  \\ \hline
Optimizer                   & AdamW  \\
LR Scheduler                & cosine \\
Learning Rate (LR)          & 0.0002                     \\
Weight Decay                & 0.01                       \\
Warm Up Steps               & 10                         \\
Micro Batch Size            & 1                          \\
Gradient Accumulation Steps & 4                          \\
QLoRA R                     & 64                         \\
QLoRA Alpha                 & 32                         \\
QLoRA Dropout               & 0.05                       \\
QLoRA Linear                & TRUE   \\
Quantization                & 4-bit 
\end{tabular}
\label{tab:hyperparameters}
\end{table}

We train the LLM in a supervised fine-tuning strategy. The loss is calculated only on the answer tokens (see Figure \ref{fig:example}). A common technique in large language model training is to apply \textit{Reinforcement Learning from Human Feedback (RLHF)} \cite{ouyang2022traininglanguagemodelsfollow}. Our hypothesis is that supervised fine-tuning is sufficient for the competition. Many answers are a single number or a list of numbers, which have an exact solutions and does not require human preferences between multiple possible answers.

\subsection{Ensemble Adapters}
Our five track solutions are created from 4 fine-tuned LoRA adapters. We call them v7, v8, v7b, and v9b. First for tracks 1,3,5 we merged v8 to base model Qwen2-72B with 56\% weight explained in Section \ref{sec:44}. For tracks 2,4 we merged v7 to base with 100\%. Version 7 adapter was trained with 417k samples whereas v8 was trained with 462k. See Table \ref{tab:training_dataset} for details about train data.

Next we trained two more LoRA adapters named v7b and v9b using two different new subsets of 152k and 40k samples respectively. Our final solutions with ensemble weights and leaderboard scores to tracks 1-5 are shown in Table \ref{tab:ensemble}.

\subsection{Wise-ft}
\label{sec:44}

To take into account the distribution shift between the evaluation data from the ShopBench dataset and the collected training data (see Section \ref{sec:datasets}), we used wise-ft \cite{wise-ft}.

Wise-ft interpolates between the weights $W_{base}$ of a base model and the weights $W_{ft}$ of a fine-tuned model using the following formula:
$$W_{wise} = (1-\alpha) * W_{base} + \alpha * W_{ft}$$
where $\alpha \in [0,1]$. This approach effectively balances the trade-off between the zero-shot capabilities of the base model ($\alpha=0$) and the task-specific performance of the fine-tuned model ($\alpha=1$). 

For LoRA, as $W_{ft} = W_{base} + W_A \cdot W_B$, we can rewrite the formula as:
 $$W_{wise} = W_{base} + \alpha * W_A \cdot W_B $$

\begin{figure}[h!]
    \centering
    \includegraphics[width=0.5\textwidth]{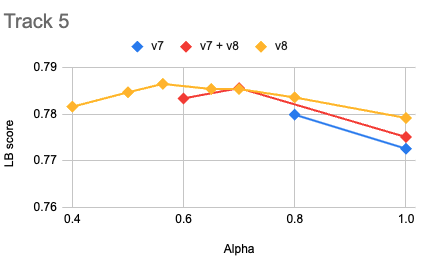}
    \caption{Accuracy gain on Track 5 using various $\alpha$ for wise-ft for 3 different versions of LoRA adapters (v7, v8 and the ensemble v7 + v8)}
    \label{fig:wise_ft}
\end{figure}

 We implemented wise-ft by rescaling the ensembled adapter weights $W_A$  and $W_B$ by a factor $\sqrt{\alpha}$, so that $(\sqrt{\alpha} * W_A) \cdot (\sqrt{\alpha} * W_B) = \alpha * W_A \cdot W_B$. For each track, we optimized $\alpha$ based on leaderboard results and obtained significant improvements for Track 1, Track 3 and Track 5 (see Figure \ref{fig:wise_ft}).

\subsection{Logits Processors}
We employed a variety of logits processors to generate outputs in specific formats. For multiple choice, ranking, and retrieval questions, we constrained our models to produce only digits and commas. For NER tasks, we enhanced the logits of the prompt tokens, encouraging the model to cite directly from the prompt. These logits processors were particularly useful in Phase 1 when we utilized less powerful models. These constrains also were useful in case when the training dataset includes only few task types. For example, you can finetune a model for MC tasks only and successfully apply it for Retrival or Reranking tasks. However, their importance diminished in Phase 2 as we transitioned to larger models that more effectively followed instructions.

\subsection{Quantization / vLLM}
KDD Cup 2024 was a code competition meaning that we must submit code plus model weights to be run on the host's pre-defined compute resources. Each participant could submit (to each track) a GitLab repository of maximize size 100GB to be executed on 4x NVIDIA T4 GPU each with 16 GB GPU memory within a time constraint.

The Qwen2-72B model is about 150GB at fp16. Therefore in order to fit this into disk and memory size constraints, we used 4bit quantization which reduced its size to 40GB.

Quantization plus using the library vLLM \cite{kwon2023efficient} accelerated our inference which allowed our model to answer all the questions within the time limit. Tracks 1-5 had 6102, 1896, 2373, 1349, and 11720 questions to be answered in 70, 20, 30, 20, 140 minutes respectively.

We improved AWQ quantization accuracy by calibrating with the 96 development questions. We compared AWQ versus GPTQ quantization and found both to be about equal in speed and accuracy.

AWQ quantization for Qwen2-72B takes about 1.5 hours on 1xA100 GPU to process. In order for the AWQ quantized Qwen2-72B to work with vLLM, we needed to pad the unquantized model with zeros to change the shape of the weights before quantization. 

\section{Results}
\label{sec:results}

\begin{table}[]
\caption{Leaderboard Results: Final is the sum of the ranks per track (lower is better). T1 - T5 are the scores per track (higher is better)}
\begin{tabular}{lrrrrrr}
\textbf{Team}              & \multicolumn{1}{l}{\textbf{Final}} & \multicolumn{1}{l}{\textbf{T1}} & \multicolumn{1}{l}{\textbf{T2}} & \multicolumn{1}{l}{\textbf{T3}} & \multicolumn{1}{l}{\textbf{T4}} & \multicolumn{1}{l}{\textbf{T5}} \\ \cline{1-7}
Team\_NVIDIA               & 5                                        & 0.833                           & 0.791                           & 0.746                           & 0.761                           & 0.788                           \\
AML\_LabCityU              & 13                                       & 0.825                           & 0.781                           & 0.728                           & 0.715                           & 0.782                           \\
shimmering\_as\_... & 18                                       & 0.824                           & 0.747                           & 0.713                           & 0.735                           & 0.763                           \\
CM\_RLLM                   & 29                                       & 0.823                           & 0.728                           & 0.722                           & 0.690                           & 0.773                           \\
ZJU\_AI4H                  & 33                                       & 0.791                           & 0.784                           & 0.694                           & 0.706                           & 0.746                           \\
BMI\_DLUT                  & \multicolumn{1}{l}{}                     & \multicolumn{1}{l}{}            & \multicolumn{1}{l}{}            & 0.733                           & \multicolumn{1}{l}{}            & \multicolumn{1}{l}{}           
\end{tabular}
\label{tab:leaderboard}
\end{table}

Our quantized, fine-tuned Qwen2-72B model achieves the highest score on each individual track (T1 - T4) and overall track T5 with a significant lead of 0.007 to 0.026 to the 2nd place (Table \ref{tab:leaderboard}). As we placed 1st in each individual track, our final score is 5, the sum of our positions, which is the highest possible score. 

Each submission for the individual track is based on the key concepts of fine-tuning a Qwen2-72B model on our developed training dataset and optionally, ensemble multiple versions and/or apply \textit{wise-tf}. The submission might differ slightly in the fine-tuning time, exact amount of training dataset and ensemble combination.

\begin{table}[]
\caption{Comparison of different base model without fine-tuning on Track 1 - Track 5. We report only scores we could run during the competition.}
\begin{tabular}{lrrrrr}
\textbf{Model} & \multicolumn{1}{l}{\textbf{T 1}} & \multicolumn{1}{l}{\textbf{T 2}} & \multicolumn{1}{l}{\textbf{T 3}} & \multicolumn{1}{l}{\textbf{T 4}} & \multicolumn{1}{l}{\textbf{T 5}} \\ \hline
Bagel-34B-v0.5 \cite{bagel34b} & 0.7007                               & 0.6609                               & 0.6339                               & 0.5871                               & 0.6834                               \\
LLaMa3-70B \cite{llama3modelcard}     & 0.7806                               & 0.6532                               & 0.6658                               & 0.6237                               & 0.7183                               \\
Smaug-72B \cite{pal2024smaug}     & 0.7178                               & \multicolumn{1}{l}{}                 & 0.6564                               & 0.6484                               & 0.6975                               \\
Qwen2-72B  \cite{yang2024qwen2technicalreport}    & 0.7982                               & 0.6407                               & 0.7193                               & 0.6918                               & 0.7486                              
\end{tabular}
\label{tab:foundation_models}
\end{table}

\begin{table}[]
\caption{Comparison foundation model as zero-shot (SZ) with fine-tuned version (FT) on Track 1 - Track 5. We report only scores we could run during the competition.}
\begin{tabular}{lrrrrr}
\textbf{Model} & \multicolumn{1}{l}{\textbf{T 1}} & \multicolumn{1}{l}{\textbf{T 2}} & \multicolumn{1}{l}{\textbf{T 3}} & \multicolumn{1}{l}{\textbf{T 4}} & \multicolumn{1}{l}{\textbf{T 5}} \\ \hline
Smaug 72B ZS   & 0.7178                           & \multicolumn{1}{l}{}             & 0.6564                           & 0.6484                           & 0.6975                           \\
Smaug 72B FT   & 0.7800                           & 0.7389                           & \multicolumn{1}{l}{}             & \multicolumn{1}{l}{}             & \multicolumn{1}{l}{}             \\
Qwen2 72B ZS   & 0.7982                           & 0.6407                           & 0.7193                           & 0.6918                           & 0.7486                           \\
Qwen2 72B FT   & 0.8334                           & 0.7909                           & 0.7461                           & 0.7609                           & 0.7883                          
\end{tabular}
\label{tab:sz_ft}
\end{table}

We provide an ablation study in Table \ref{tab:foundation_models}, \ref{tab:sz_ft} and \ref{tab:ensemble}. Some values are missing in the tables due to failed submissions and the successful submissions were sufficient to decide the next experiments. First, table \ref{tab:foundation_models} compares different base models without being fine-tuned. We observe that Qwen2-72B has the highest score except of for Track 2, followed by LLaMa3-70B is 2nd place except of Track 4. The performance of the models are equivalent to public LLMs benchmarks.

Next, we fine-tuned Smaug-72B and Qwen2-72B on our training dataset. We compare the zero-shot version (SZ) with the fine-tuned version (FT) as seen in Table \ref{tab:sz_ft}. Qwen2-72B SZ would scored 4th place on Track 5, demonstrating that base model provide great capabilities without fine-tuning. We achieve significant gains of 0.0035 to 0.15 by additional fine-tuning.

\begin{table}[]
\caption{Configuration and results of ensembling multiple LoRA configurations. B is the base model weights. We add the weights M1 from the LoRA Adapter multiplied with the weight W1 (equivalent for the ensemble B+W1xM1+W2xM2). LB demonstrates the scores on the leaderboard.}
\begin{tabular}{lccccc}
\textbf{Model}                        & \textbf{T 1}    & \textbf{T 2}    & \textbf{T 3}    & \textbf{T 4}    & \textbf{T 5}    \\ \hline
LoRA 1 name (M1)              & v8              & v7              & v8              & v7              & v8              \\
LoRA 1 weight (W1)            & 0.56            & 1.0             & 0.56            & 1.0             & 0.56            \\
LB B+W1xM1          & 0.831          & 0.787          & 0.742          & 0.758          & 0.787          \\ \hline
LoRA 2 name (M2)              & v9b             & v7b             & v9b             & v7b             & v9b             \\
LoRA 2 weight (W2)            & 0.75            & 0.5             & 0.25            & 0.5             & 0.25            \\
\textbf{LB B+W1xM1+W2xM2} & \textbf{0.833} & \textbf{0.791} & \textbf{0.746} & \textbf{0.761} & \textbf{0.788}
\end{tabular}
\label{tab:ensemble}
\end{table}

Table \ref{tab:ensemble} summarize the effect of ensembling multiple models. The first solution is to submit the base model merged with the first LoRA adapter, scaled by weight W1. If we ensemble two adapters (\textit{LB B+W1xM1+W2xM2}), we observe that the LB score improves between 0.001 to 0.004.

\section{Conclusion}
The KDD Cup 2024 was a great competition with a diverse set of tasks to evaluate Large Language Models capabilities in the domain of online shopping. The code competition design ensured a fair comparison of solutions. Our team solution is a single models with multiple optimization methods, which scored the 1st place on each track. It was essential to fine-tune a base model with an additional training dataset. The lack of an official training dataset was compensated by processing multiple public datasets and prompting Large Language Models. We ensembled multiple LoRA adapater, applied \textit{wise-ft} for distribution shift and constrained the model output with a \textit{Logits Processors}. We optimized inference with 4-bit quantization and vLLM to run a 72 billion parameters model on 4x NVIDIA T4 with each 16 GB GPU memory in the time constrain. In addition, we share multiple experiments as an ablation study.

\bibliographystyle{ACM-Reference-Format}
\bibliography{sample}


\begin{thebibliography}{18}


\ifx \showCODEN    \undefined \def \showCODEN     #1{\unskip}     \fi
\ifx \showDOI      \undefined \def \showDOI       #1{#1}\fi
\ifx \showISBNx    \undefined \def \showISBNx     #1{\unskip}     \fi
\ifx \showISBNxiii \undefined \def \showISBNxiii  #1{\unskip}     \fi
\ifx \showISSN     \undefined \def \showISSN      #1{\unskip}     \fi
\ifx \showLCCN     \undefined \def \showLCCN      #1{\unskip}     \fi
\ifx \shownote     \undefined \def \shownote      #1{#1}          \fi
\ifx \showarticletitle \undefined \def \showarticletitle #1{#1}   \fi
\ifx \showURL      \undefined \def \showURL       {\relax}        \fi
\providecommand\bibfield[2]{#2}
\providecommand\bibinfo[2]{#2}
\providecommand\natexlab[1]{#1}
\providecommand\showeprint[2][]{arXiv:#2}

\bibitem[bag(2024)]%
        {bagel34b}
 \bibinfo{year}{2024}\natexlab{}.
\newblock \showarticletitle{A bagel, with everything (except DPO)}.
\newblock  (\bibinfo{year}{2024}).
\newblock
\urldef\tempurl%
\url{https://huggingface.co/jondurbin/bagel-34b-v0.5}
\showURL{%
\tempurl}


\bibitem[kdd(2024)]%
        {kddcup2024}
 \bibinfo{year}{2024}\natexlab{}.
\newblock \showarticletitle{A Multi-task Online Shopping Challenge for Large Language Models}.
\newblock  (\bibinfo{year}{2024}).
\newblock
\urldef\tempurl%
\url{https://amazon-kddcup24.github.io/}
\showURL{%
\tempurl}


\bibitem[AI@Meta(2024)]%
        {llama3modelcard}
\bibfield{author}{\bibinfo{person}{AI@Meta}.} \bibinfo{year}{2024}\natexlab{}.
\newblock \showarticletitle{Llama 3 Model Card}.
\newblock  (\bibinfo{year}{2024}).
\newblock
\urldef\tempurl%
\url{https://github.com/meta-llama/llama3/blob/main/MODEL_CARD.md}
\showURL{%
\tempurl}


\bibitem[Burgess et~al\mbox{.}(2019)]%
        {8877390}
\bibfield{author}{\bibinfo{person}{Neil Burgess}, \bibinfo{person}{Jelena Milanovic}, \bibinfo{person}{Nigel Stephens}, \bibinfo{person}{Konstantinos Monachopoulos}, {and} \bibinfo{person}{David Mansell}.} \bibinfo{year}{2019}\natexlab{}.
\newblock \showarticletitle{Bfloat16 Processing for Neural Networks}. In \bibinfo{booktitle}{\emph{2019 IEEE 26th Symposium on Computer Arithmetic (ARITH)}}. \bibinfo{pages}{88--91}.
\newblock
\urldef\tempurl%
\url{https://doi.org/10.1109/ARITH.2019.00022}
\showDOI{\tempurl}


\bibitem[Dettmers et~al\mbox{.}(2023)]%
        {dettmers2023qloraefficientfinetuningquantized}
\bibfield{author}{\bibinfo{person}{Tim Dettmers}, \bibinfo{person}{Artidoro Pagnoni}, \bibinfo{person}{Ari Holtzman}, {and} \bibinfo{person}{Luke Zettlemoyer}.} \bibinfo{year}{2023}\natexlab{}.
\newblock \bibinfo{title}{QLoRA: Efficient Finetuning of Quantized LLMs}.
\newblock
\newblock
\showeprint[arxiv]{2305.14314}~[cs.LG]
\urldef\tempurl%
\url{https://arxiv.org/abs/2305.14314}
\showURL{%
\tempurl}


\bibitem[Hendrycks et~al\mbox{.}(2021a)]%
        {hendrycks2021ethics}
\bibfield{author}{\bibinfo{person}{Dan Hendrycks}, \bibinfo{person}{Collin Burns}, \bibinfo{person}{Steven Basart}, \bibinfo{person}{Andrew Critch}, \bibinfo{person}{Jerry Li}, \bibinfo{person}{Dawn Song}, {and} \bibinfo{person}{Jacob Steinhardt}.} \bibinfo{year}{2021}\natexlab{a}.
\newblock \showarticletitle{Aligning AI With Shared Human Values}.
\newblock \bibinfo{journal}{\emph{Proceedings of the International Conference on Learning Representations (ICLR)}} (\bibinfo{year}{2021}).
\newblock


\bibitem[Hendrycks et~al\mbox{.}(2021b)]%
        {hendryckstest2021}
\bibfield{author}{\bibinfo{person}{Dan Hendrycks}, \bibinfo{person}{Collin Burns}, \bibinfo{person}{Steven Basart}, \bibinfo{person}{Andy Zou}, \bibinfo{person}{Mantas Mazeika}, \bibinfo{person}{Dawn Song}, {and} \bibinfo{person}{Jacob Steinhardt}.} \bibinfo{year}{2021}\natexlab{b}.
\newblock \showarticletitle{Measuring Massive Multitask Language Understanding}.
\newblock \bibinfo{journal}{\emph{Proceedings of the International Conference on Learning Representations (ICLR)}} (\bibinfo{year}{2021}).
\newblock


\bibitem[Hou et~al\mbox{.}(2024)]%
        {hou2024bridging}
\bibfield{author}{\bibinfo{person}{Yupeng Hou}, \bibinfo{person}{Jiacheng Li}, \bibinfo{person}{Zhankui He}, \bibinfo{person}{An Yan}, \bibinfo{person}{Xiusi Chen}, {and} \bibinfo{person}{Julian McAuley}.} \bibinfo{year}{2024}\natexlab{}.
\newblock \showarticletitle{Bridging Language and Items for Retrieval and Recommendation}.
\newblock \bibinfo{journal}{\emph{arXiv preprint arXiv:2403.03952}} (\bibinfo{year}{2024}).
\newblock


\bibitem[Jin et~al\mbox{.}(2023)]%
        {jin2023amazonm2multilingualmultilocaleshopping}
\bibfield{author}{\bibinfo{person}{Wei Jin}, \bibinfo{person}{Haitao Mao}, \bibinfo{person}{Zheng Li}, \bibinfo{person}{Haoming Jiang}, \bibinfo{person}{Chen Luo}, \bibinfo{person}{Hongzhi Wen}, \bibinfo{person}{Haoyu Han}, \bibinfo{person}{Hanqing Lu}, \bibinfo{person}{Zhengyang Wang}, \bibinfo{person}{Ruirui Li}, \bibinfo{person}{Zhen Li}, \bibinfo{person}{Monica~Xiao Cheng}, \bibinfo{person}{Rahul Goutam}, \bibinfo{person}{Haiyang Zhang}, \bibinfo{person}{Karthik Subbian}, \bibinfo{person}{Suhang Wang}, \bibinfo{person}{Yizhou Sun}, \bibinfo{person}{Jiliang Tang}, \bibinfo{person}{Bing Yin}, {and} \bibinfo{person}{Xianfeng Tang}.} \bibinfo{year}{2023}\natexlab{}.
\newblock \bibinfo{title}{Amazon-M2: A Multilingual Multi-locale Shopping Session Dataset for Recommendation and Text Generation}.
\newblock
\newblock
\showeprint[arxiv]{2307.09688}~[cs.IR]
\urldef\tempurl%
\url{https://arxiv.org/abs/2307.09688}
\showURL{%
\tempurl}


\bibitem[Kwon et~al\mbox{.}(2023)]%
        {kwon2023efficient}
\bibfield{author}{\bibinfo{person}{Woosuk Kwon}, \bibinfo{person}{Zhuohan Li}, \bibinfo{person}{Siyuan Zhuang}, \bibinfo{person}{Ying Sheng}, \bibinfo{person}{Lianmin Zheng}, \bibinfo{person}{Cody~Hao Yu}, \bibinfo{person}{Joseph~E. Gonzalez}, \bibinfo{person}{Hao Zhang}, {and} \bibinfo{person}{Ion Stoica}.} \bibinfo{year}{2023}\natexlab{}.
\newblock \showarticletitle{Efficient Memory Management for Large Language Model Serving with PagedAttention}. In \bibinfo{booktitle}{\emph{Proceedings of the ACM SIGOPS 29th Symposium on Operating Systems Principles}}.
\newblock


\bibitem[OpenAI et~al\mbox{.}(2024)]%
        {openai2024gpt4technicalreport}
\bibfield{author}{\bibinfo{person}{OpenAI}, \bibinfo{person}{Josh Achiam}, \bibinfo{person}{Steven Adler}, \bibinfo{person}{Sandhini Agarwal}, \bibinfo{person}{Lama Ahmad}, \bibinfo{person}{Ilge Akkaya}, \bibinfo{person}{Florencia~Leoni Aleman}, \bibinfo{person}{Diogo Almeida}, \bibinfo{person}{Janko Altenschmidt}, \bibinfo{person}{Sam Altman}, \bibinfo{person}{Shyamal Anadkat}, \bibinfo{person}{Red Avila}, \bibinfo{person}{Igor Babuschkin}, \bibinfo{person}{Suchir Balaji}, \bibinfo{person}{Valerie Balcom}, \bibinfo{person}{Paul Baltescu}, \bibinfo{person}{Haiming Bao}, \bibinfo{person}{Mohammad Bavarian}, \bibinfo{person}{Jeff Belgum}, \bibinfo{person}{Irwan Bello}, \bibinfo{person}{Jake Berdine}, \bibinfo{person}{Gabriel Bernadett-Shapiro}, \bibinfo{person}{Christopher Berner}, \bibinfo{person}{Lenny Bogdonoff}, \bibinfo{person}{Oleg Boiko}, \bibinfo{person}{Madelaine Boyd}, \bibinfo{person}{Anna-Luisa Brakman}, \bibinfo{person}{Greg Brockman}, \bibinfo{person}{Tim Brooks}, \bibinfo{person}{Miles Brundage},
  \bibinfo{person}{Kevin Button}, \bibinfo{person}{Trevor Cai}, \bibinfo{person}{Rosie Campbell}, \bibinfo{person}{Andrew Cann}, \bibinfo{person}{Brittany Carey}, \bibinfo{person}{Chelsea Carlson}, \bibinfo{person}{Rory Carmichael}, \bibinfo{person}{Brooke Chan}, \bibinfo{person}{Che Chang}, \bibinfo{person}{Fotis Chantzis}, \bibinfo{person}{Derek Chen}, \bibinfo{person}{Sully Chen}, \bibinfo{person}{Ruby Chen}, \bibinfo{person}{Jason Chen}, \bibinfo{person}{Mark Chen}, \bibinfo{person}{Ben Chess}, \bibinfo{person}{Chester Cho}, \bibinfo{person}{Casey Chu}, \bibinfo{person}{Hyung~Won Chung}, \bibinfo{person}{Dave Cummings}, \bibinfo{person}{Jeremiah Currier}, \bibinfo{person}{Yunxing Dai}, \bibinfo{person}{Cory Decareaux}, \bibinfo{person}{Thomas Degry}, \bibinfo{person}{Noah Deutsch}, \bibinfo{person}{Damien Deville}, \bibinfo{person}{Arka Dhar}, \bibinfo{person}{David Dohan}, \bibinfo{person}{Steve Dowling}, \bibinfo{person}{Sheila Dunning}, \bibinfo{person}{Adrien Ecoffet}, \bibinfo{person}{Atty Eleti},
  \bibinfo{person}{Tyna Eloundou}, \bibinfo{person}{David Farhi}, \bibinfo{person}{Liam Fedus}, \bibinfo{person}{Niko Felix}, \bibinfo{person}{Simón~Posada Fishman}, \bibinfo{person}{Juston Forte}, \bibinfo{person}{Isabella Fulford}, \bibinfo{person}{Leo Gao}, \bibinfo{person}{Elie Georges}, \bibinfo{person}{Christian Gibson}, \bibinfo{person}{Vik Goel}, \bibinfo{person}{Tarun Gogineni}, \bibinfo{person}{Gabriel Goh}, \bibinfo{person}{Rapha Gontijo-Lopes}, \bibinfo{person}{Jonathan Gordon}, \bibinfo{person}{Morgan Grafstein}, \bibinfo{person}{Scott Gray}, \bibinfo{person}{Ryan Greene}, \bibinfo{person}{Joshua Gross}, \bibinfo{person}{Shixiang~Shane Gu}, \bibinfo{person}{Yufei Guo}, \bibinfo{person}{Chris Hallacy}, \bibinfo{person}{Jesse Han}, \bibinfo{person}{Jeff Harris}, \bibinfo{person}{Yuchen He}, \bibinfo{person}{Mike Heaton}, \bibinfo{person}{Johannes Heidecke}, \bibinfo{person}{Chris Hesse}, \bibinfo{person}{Alan Hickey}, \bibinfo{person}{Wade Hickey}, \bibinfo{person}{Peter Hoeschele},
  \bibinfo{person}{Brandon Houghton}, \bibinfo{person}{Kenny Hsu}, \bibinfo{person}{Shengli Hu}, \bibinfo{person}{Xin Hu}, \bibinfo{person}{Joost Huizinga}, \bibinfo{person}{Shantanu Jain}, \bibinfo{person}{Shawn Jain}, \bibinfo{person}{Joanne Jang}, \bibinfo{person}{Angela Jiang}, \bibinfo{person}{Roger Jiang}, \bibinfo{person}{Haozhun Jin}, \bibinfo{person}{Denny Jin}, \bibinfo{person}{Shino Jomoto}, \bibinfo{person}{Billie Jonn}, \bibinfo{person}{Heewoo Jun}, \bibinfo{person}{Tomer Kaftan}, \bibinfo{person}{Łukasz Kaiser}, \bibinfo{person}{Ali Kamali}, \bibinfo{person}{Ingmar Kanitscheider}, \bibinfo{person}{Nitish~Shirish Keskar}, \bibinfo{person}{Tabarak Khan}, \bibinfo{person}{Logan Kilpatrick}, \bibinfo{person}{Jong~Wook Kim}, \bibinfo{person}{Christina Kim}, \bibinfo{person}{Yongjik Kim}, \bibinfo{person}{Jan~Hendrik Kirchner}, \bibinfo{person}{Jamie Kiros}, \bibinfo{person}{Matt Knight}, \bibinfo{person}{Daniel Kokotajlo}, \bibinfo{person}{Łukasz Kondraciuk}, \bibinfo{person}{Andrew Kondrich},
  \bibinfo{person}{Aris Konstantinidis}, \bibinfo{person}{Kyle Kosic}, \bibinfo{person}{Gretchen Krueger}, \bibinfo{person}{Vishal Kuo}, \bibinfo{person}{Michael Lampe}, \bibinfo{person}{Ikai Lan}, \bibinfo{person}{Teddy Lee}, \bibinfo{person}{Jan Leike}, \bibinfo{person}{Jade Leung}, \bibinfo{person}{Daniel Levy}, \bibinfo{person}{Chak~Ming Li}, \bibinfo{person}{Rachel Lim}, \bibinfo{person}{Molly Lin}, \bibinfo{person}{Stephanie Lin}, \bibinfo{person}{Mateusz Litwin}, \bibinfo{person}{Theresa Lopez}, \bibinfo{person}{Ryan Lowe}, \bibinfo{person}{Patricia Lue}, \bibinfo{person}{Anna Makanju}, \bibinfo{person}{Kim Malfacini}, \bibinfo{person}{Sam Manning}, \bibinfo{person}{Todor Markov}, \bibinfo{person}{Yaniv Markovski}, \bibinfo{person}{Bianca Martin}, \bibinfo{person}{Katie Mayer}, \bibinfo{person}{Andrew Mayne}, \bibinfo{person}{Bob McGrew}, \bibinfo{person}{Scott~Mayer McKinney}, \bibinfo{person}{Christine McLeavey}, \bibinfo{person}{Paul McMillan}, \bibinfo{person}{Jake McNeil}, \bibinfo{person}{David
  Medina}, \bibinfo{person}{Aalok Mehta}, \bibinfo{person}{Jacob Menick}, \bibinfo{person}{Luke Metz}, \bibinfo{person}{Andrey Mishchenko}, \bibinfo{person}{Pamela Mishkin}, \bibinfo{person}{Vinnie Monaco}, \bibinfo{person}{Evan Morikawa}, \bibinfo{person}{Daniel Mossing}, \bibinfo{person}{Tong Mu}, \bibinfo{person}{Mira Murati}, \bibinfo{person}{Oleg Murk}, \bibinfo{person}{David Mély}, \bibinfo{person}{Ashvin Nair}, \bibinfo{person}{Reiichiro Nakano}, \bibinfo{person}{Rajeev Nayak}, \bibinfo{person}{Arvind Neelakantan}, \bibinfo{person}{Richard Ngo}, \bibinfo{person}{Hyeonwoo Noh}, \bibinfo{person}{Long Ouyang}, \bibinfo{person}{Cullen O'Keefe}, \bibinfo{person}{Jakub Pachocki}, \bibinfo{person}{Alex Paino}, \bibinfo{person}{Joe Palermo}, \bibinfo{person}{Ashley Pantuliano}, \bibinfo{person}{Giambattista Parascandolo}, \bibinfo{person}{Joel Parish}, \bibinfo{person}{Emy Parparita}, \bibinfo{person}{Alex Passos}, \bibinfo{person}{Mikhail Pavlov}, \bibinfo{person}{Andrew Peng}, \bibinfo{person}{Adam
  Perelman}, \bibinfo{person}{Filipe de Avila Belbute~Peres}, \bibinfo{person}{Michael Petrov}, \bibinfo{person}{Henrique~Ponde de Oliveira~Pinto}, \bibinfo{person}{Michael}, \bibinfo{person}{Pokorny}, \bibinfo{person}{Michelle Pokrass}, \bibinfo{person}{Vitchyr~H. Pong}, \bibinfo{person}{Tolly Powell}, \bibinfo{person}{Alethea Power}, \bibinfo{person}{Boris Power}, \bibinfo{person}{Elizabeth Proehl}, \bibinfo{person}{Raul Puri}, \bibinfo{person}{Alec Radford}, \bibinfo{person}{Jack Rae}, \bibinfo{person}{Aditya Ramesh}, \bibinfo{person}{Cameron Raymond}, \bibinfo{person}{Francis Real}, \bibinfo{person}{Kendra Rimbach}, \bibinfo{person}{Carl Ross}, \bibinfo{person}{Bob Rotsted}, \bibinfo{person}{Henri Roussez}, \bibinfo{person}{Nick Ryder}, \bibinfo{person}{Mario Saltarelli}, \bibinfo{person}{Ted Sanders}, \bibinfo{person}{Shibani Santurkar}, \bibinfo{person}{Girish Sastry}, \bibinfo{person}{Heather Schmidt}, \bibinfo{person}{David Schnurr}, \bibinfo{person}{John Schulman}, \bibinfo{person}{Daniel Selsam},
  \bibinfo{person}{Kyla Sheppard}, \bibinfo{person}{Toki Sherbakov}, \bibinfo{person}{Jessica Shieh}, \bibinfo{person}{Sarah Shoker}, \bibinfo{person}{Pranav Shyam}, \bibinfo{person}{Szymon Sidor}, \bibinfo{person}{Eric Sigler}, \bibinfo{person}{Maddie Simens}, \bibinfo{person}{Jordan Sitkin}, \bibinfo{person}{Katarina Slama}, \bibinfo{person}{Ian Sohl}, \bibinfo{person}{Benjamin Sokolowsky}, \bibinfo{person}{Yang Song}, \bibinfo{person}{Natalie Staudacher}, \bibinfo{person}{Felipe~Petroski Such}, \bibinfo{person}{Natalie Summers}, \bibinfo{person}{Ilya Sutskever}, \bibinfo{person}{Jie Tang}, \bibinfo{person}{Nikolas Tezak}, \bibinfo{person}{Madeleine~B. Thompson}, \bibinfo{person}{Phil Tillet}, \bibinfo{person}{Amin Tootoonchian}, \bibinfo{person}{Elizabeth Tseng}, \bibinfo{person}{Preston Tuggle}, \bibinfo{person}{Nick Turley}, \bibinfo{person}{Jerry Tworek}, \bibinfo{person}{Juan Felipe~Cerón Uribe}, \bibinfo{person}{Andrea Vallone}, \bibinfo{person}{Arun Vijayvergiya}, \bibinfo{person}{Chelsea Voss},
  \bibinfo{person}{Carroll Wainwright}, \bibinfo{person}{Justin~Jay Wang}, \bibinfo{person}{Alvin Wang}, \bibinfo{person}{Ben Wang}, \bibinfo{person}{Jonathan Ward}, \bibinfo{person}{Jason Wei}, \bibinfo{person}{CJ Weinmann}, \bibinfo{person}{Akila Welihinda}, \bibinfo{person}{Peter Welinder}, \bibinfo{person}{Jiayi Weng}, \bibinfo{person}{Lilian Weng}, \bibinfo{person}{Matt Wiethoff}, \bibinfo{person}{Dave Willner}, \bibinfo{person}{Clemens Winter}, \bibinfo{person}{Samuel Wolrich}, \bibinfo{person}{Hannah Wong}, \bibinfo{person}{Lauren Workman}, \bibinfo{person}{Sherwin Wu}, \bibinfo{person}{Jeff Wu}, \bibinfo{person}{Michael Wu}, \bibinfo{person}{Kai Xiao}, \bibinfo{person}{Tao Xu}, \bibinfo{person}{Sarah Yoo}, \bibinfo{person}{Kevin Yu}, \bibinfo{person}{Qiming Yuan}, \bibinfo{person}{Wojciech Zaremba}, \bibinfo{person}{Rowan Zellers}, \bibinfo{person}{Chong Zhang}, \bibinfo{person}{Marvin Zhang}, \bibinfo{person}{Shengjia Zhao}, \bibinfo{person}{Tianhao Zheng}, \bibinfo{person}{Juntang Zhuang},
  \bibinfo{person}{William Zhuk}, {and} \bibinfo{person}{Barret Zoph}.} \bibinfo{year}{2024}\natexlab{}.
\newblock \bibinfo{title}{GPT-4 Technical Report}.
\newblock
\newblock
\showeprint[arxiv]{2303.08774}~[cs.CL]
\urldef\tempurl%
\url{https://arxiv.org/abs/2303.08774}
\showURL{%
\tempurl}


\bibitem[Ouyang et~al\mbox{.}(2022)]%
        {ouyang2022traininglanguagemodelsfollow}
\bibfield{author}{\bibinfo{person}{Long Ouyang}, \bibinfo{person}{Jeff Wu}, \bibinfo{person}{Xu Jiang}, \bibinfo{person}{Diogo Almeida}, \bibinfo{person}{Carroll~L. Wainwright}, \bibinfo{person}{Pamela Mishkin}, \bibinfo{person}{Chong Zhang}, \bibinfo{person}{Sandhini Agarwal}, \bibinfo{person}{Katarina Slama}, \bibinfo{person}{Alex Ray}, \bibinfo{person}{John Schulman}, \bibinfo{person}{Jacob Hilton}, \bibinfo{person}{Fraser Kelton}, \bibinfo{person}{Luke Miller}, \bibinfo{person}{Maddie Simens}, \bibinfo{person}{Amanda Askell}, \bibinfo{person}{Peter Welinder}, \bibinfo{person}{Paul Christiano}, \bibinfo{person}{Jan Leike}, {and} \bibinfo{person}{Ryan Lowe}.} \bibinfo{year}{2022}\natexlab{}.
\newblock \bibinfo{title}{Training language models to follow instructions with human feedback}.
\newblock
\newblock
\showeprint[arxiv]{2203.02155}~[cs.CL]
\urldef\tempurl%
\url{https://arxiv.org/abs/2203.02155}
\showURL{%
\tempurl}


\bibitem[Pal et~al\mbox{.}(2024)]%
        {pal2024smaug}
\bibfield{author}{\bibinfo{person}{Arka Pal}, \bibinfo{person}{Deep Karkhanis}, \bibinfo{person}{Samuel Dooley}, \bibinfo{person}{Manley Roberts}, \bibinfo{person}{Siddartha Naidu}, {and} \bibinfo{person}{Colin White}.} \bibinfo{year}{2024}\natexlab{}.
\newblock \showarticletitle{Smaug: Fixing Failure Modes of Preference Optimisation with DPO-Positive}.
\newblock \bibinfo{journal}{\emph{arXiv preprint arXiv:2402.13228}} (\bibinfo{year}{2024}).
\newblock


\bibitem[Peng et~al\mbox{.}(2024)]%
        {peng2024ecellm}
\bibfield{author}{\bibinfo{person}{Bo Peng}, \bibinfo{person}{Xinyi Ling}, \bibinfo{person}{Ziru Chen}, \bibinfo{person}{Huan Sun}, {and} \bibinfo{person}{Xia Ning}.} \bibinfo{year}{2024}\natexlab{}.
\newblock \showarticletitle{eCe{LLM}: Generalizing Large Language Models for E-commerce from Large-scale, High-quality Instruction Data}. In \bibinfo{booktitle}{\emph{Forty-first International Conference on Machine Learning}}.
\newblock
\urldef\tempurl%
\url{https://openreview.net/forum?id=LWRI4uPG2X}
\showURL{%
\tempurl}


\bibitem[Reddy et~al\mbox{.}(2022)]%
        {reddy2022shopping}
\bibfield{author}{\bibinfo{person}{Chandan~K. Reddy}, \bibinfo{person}{Lluís Màrquez}, \bibinfo{person}{Fran Valero}, \bibinfo{person}{Nikhil Rao}, \bibinfo{person}{Hugo Zaragoza}, \bibinfo{person}{Sambaran Bandyopadhyay}, \bibinfo{person}{Arnab Biswas}, \bibinfo{person}{Anlu Xing}, {and} \bibinfo{person}{Karthik Subbian}.} \bibinfo{year}{2022}\natexlab{}.
\newblock \showarticletitle{Shopping Queries Dataset: A Large-Scale {ESCI} Benchmark for Improving Product Search}.
\newblock  (\bibinfo{year}{2022}).
\newblock
\showeprint[arxiv]{2206.06588}


\bibitem[Taori et~al\mbox{.}(2023)]%
        {alpaca}
\bibfield{author}{\bibinfo{person}{Rohan Taori}, \bibinfo{person}{Ishaan Gulrajani}, \bibinfo{person}{Tianyi Zhang}, \bibinfo{person}{Yann Dubois}, \bibinfo{person}{Xuechen Li}, \bibinfo{person}{Carlos Guestrin}, \bibinfo{person}{Percy Liang}, {and} \bibinfo{person}{Tatsunori~B. Hashimoto}.} \bibinfo{year}{2023}\natexlab{}.
\newblock \bibinfo{title}{Stanford Alpaca: An Instruction-following LLaMA model}.
\newblock \bibinfo{howpublished}{\url{https://github.com/tatsu-lab/stanford_alpaca}}.
\newblock


\bibitem[Wortsman et~al\mbox{.}(2021)]%
        {wise-ft}
\bibfield{author}{\bibinfo{person}{Mitchell Wortsman}, \bibinfo{person}{Gabriel Ilharco}, \bibinfo{person}{Jong~Wook Kim}, \bibinfo{person}{Mike Li}, \bibinfo{person}{Simon Kornblith}, \bibinfo{person}{Rebecca Roelofs}, \bibinfo{person}{Raphael Gontijo-Lopes}, \bibinfo{person}{Hannaneh Hajishirzi}, \bibinfo{person}{Ali Farhadi}, \bibinfo{person}{Hongseok Namkoong}, {and} \bibinfo{person}{Ludwig Schmidt}.} \bibinfo{year}{2021}\natexlab{}.
\newblock \bibinfo{title}{Robust fine-tuning of zero-shot models}.
\newblock
\newblock
\showeprint{arXiv:2109.01903}


\bibitem[Yang et~al\mbox{.}(2024)]%
        {yang2024qwen2technicalreport}
\bibfield{author}{\bibinfo{person}{An Yang}, \bibinfo{person}{Baosong Yang}, \bibinfo{person}{Binyuan Hui}, \bibinfo{person}{Bo Zheng}, \bibinfo{person}{Bowen Yu}, \bibinfo{person}{Chang Zhou}, \bibinfo{person}{Chengpeng Li}, \bibinfo{person}{Chengyuan Li}, \bibinfo{person}{Dayiheng Liu}, \bibinfo{person}{Fei Huang}, \bibinfo{person}{Guanting Dong}, \bibinfo{person}{Haoran Wei}, \bibinfo{person}{Huan Lin}, \bibinfo{person}{Jialong Tang}, \bibinfo{person}{Jialin Wang}, \bibinfo{person}{Jian Yang}, \bibinfo{person}{Jianhong Tu}, \bibinfo{person}{Jianwei Zhang}, \bibinfo{person}{Jianxin Ma}, \bibinfo{person}{Jianxin Yang}, \bibinfo{person}{Jin Xu}, \bibinfo{person}{Jingren Zhou}, \bibinfo{person}{Jinze Bai}, \bibinfo{person}{Jinzheng He}, \bibinfo{person}{Junyang Lin}, \bibinfo{person}{Kai Dang}, \bibinfo{person}{Keming Lu}, \bibinfo{person}{Keqin Chen}, \bibinfo{person}{Kexin Yang}, \bibinfo{person}{Mei Li}, \bibinfo{person}{Mingfeng Xue}, \bibinfo{person}{Na Ni}, \bibinfo{person}{Pei Zhang},
  \bibinfo{person}{Peng Wang}, \bibinfo{person}{Ru Peng}, \bibinfo{person}{Rui Men}, \bibinfo{person}{Ruize Gao}, \bibinfo{person}{Runji Lin}, \bibinfo{person}{Shijie Wang}, \bibinfo{person}{Shuai Bai}, \bibinfo{person}{Sinan Tan}, \bibinfo{person}{Tianhang Zhu}, \bibinfo{person}{Tianhao Li}, \bibinfo{person}{Tianyu Liu}, \bibinfo{person}{Wenbin Ge}, \bibinfo{person}{Xiaodong Deng}, \bibinfo{person}{Xiaohuan Zhou}, \bibinfo{person}{Xingzhang Ren}, \bibinfo{person}{Xinyu Zhang}, \bibinfo{person}{Xipin Wei}, \bibinfo{person}{Xuancheng Ren}, \bibinfo{person}{Xuejing Liu}, \bibinfo{person}{Yang Fan}, \bibinfo{person}{Yang Yao}, \bibinfo{person}{Yichang Zhang}, \bibinfo{person}{Yu Wan}, \bibinfo{person}{Yunfei Chu}, \bibinfo{person}{Yuqiong Liu}, \bibinfo{person}{Zeyu Cui}, \bibinfo{person}{Zhenru Zhang}, \bibinfo{person}{Zhifang Guo}, {and} \bibinfo{person}{Zhihao Fan}.} \bibinfo{year}{2024}\natexlab{}.
\newblock \bibinfo{title}{Qwen2 Technical Report}.
\newblock
\newblock
\showeprint[arxiv]{2407.10671}~[cs.CL]
\urldef\tempurl%
\url{https://arxiv.org/abs/2407.10671}
\showURL{%
\tempurl}


\end{thebibliography}

\appendix

\onecolumn
\section{Details on Training Dataset}
\label{sec:training_datasets}

In Table \ref{tab:training_dataset}, we provide an overview of the different datasets we generated. We share which dataset was used as a source input. We describe which task the resulting dataset is most similar to (column \textit{Task}), the size and if a LLM was used. Finally, we provide additional explanation for our own ideas.

\begin{table}[]
\tiny
\caption{Overview of the different training datasets we developed for fine-tuning Qwen2-72B. The task column describes if our processed dataset is close to an existing dataset task. The column LLM indicates if a LLM was used for generating the dataset.}
\begin{tabular}{rllllrll}
\multicolumn{1}{l}{\textbf{No}} & \textbf{Source Dataset} & \textbf{Task}                                  & \textbf{Task Type}       & \textbf{Adapter} & \multicolumn{1}{l}{\textbf{Size}} & \textbf{LLM} & \textbf{Additional Explanation}                                                                                                                                                                                                                                                                                                                                                                                                                                                   \\ \hline
1                               & Amazon-M2               & KDD Cup2024 Task 2                             & multiple-choice          & v7, v8           & 2350                              & Yes          & Select product categories given product attributes \\
2                               & Amazon Reviews 2023     & KDD Cup2024 Task 3                             & retrieval                & v7, v7b, v8      & 7373                              & Yes          & Given a product type and sentiment, select 3 most likely snippet a customer would write about the product                                                                                                                                                                                                                                                                                                                                                                         \\
3                               & Amazon Reviews 2023     & KDD Cup2024 Task 7                             & retrieval                & v7, v7b, v8      & 3608                              & Yes          & Given a product type and a review, select 3 aspects covered by the review                                                                                                                                                                                                                                                                                                                                                                                                         \\
4                               & Amazon Reviews 2023     & KDD Cup2024 Task 10                            & multiple-choice          & v7, v7b, v8      & 10000                             & Yes          & Given a product type, which of the following categories complement the product type best?                                                                                                                                                                                                                                                                                                                                                                                         \\
5                               & ESCI-data               & KDD Cup2024 Task 12                            & ranking                  & v7, v8           & 16728                             & No           &                                                                                                                                                                                                                                                                                                                                                                                                                                                                                   \\
6                               & Amazon Reviews 2023     & KDD Cup2024 Task 14                            & ranking                  & v7, v7b, v8      & 5815                              & No           & Given a product title a customer will buy, which other product titles will he like                                                                                                                                                                                                                                                                                                                                                                                                \\
7                               & Amazon Reviews 2023     & KDD Cup2024 Task 15                            & multiple-choice          & v7, v7b, v8      & 10000                             & No           & Given a product review, estimate the rating of the review                                                                                                                                                                                                                                                                                                                                                                                                                         \\
8                               & ESCI-data               & KDD Cup2024 Task 16                            & multiple-choice          & v7, v8           & 10000                             & No           &                                                                                                                                                                                                                                                                                                                                                                                                                                                                                   \\
9                               & Amazon-M2               & KDD Cup2024 Task 17                            & generation               & v7, v8           & 10000                             & No           &                                                                                                                                                                                                                                                                                                                                                                                                                                                                                   \\
10                              & Amazon-M2               & KDD Cup2024 Task 18                            & multiple-choice          & v7, v8           & 10000                             & No           &                                                                                                                                                                                                                                                                                                                                                                                                                                                                                   \\
11                              & NingLab/ECInstruct      & Attribute Value Extraction        & named entity recognition & v7, v8           & 19622                             & Yes          &                                                                                                                                                                                                                                                                                                                                                                                                                                                                                   \\
12                              & NingLab/ECInstruct      & Multiclass Product Classification & multiple-choice          & v7, v8           & 10000                             & Yes          &                                                                                                                                                                                                                                                                                                                                                                                                                                                                                   \\
13                              & NingLab/ECInstruct      & Product Relation Prediction       & multiple-choice          & v7, v8           & 10000                             & Yes          &                                                                                                                                                                                                                                                                                                                                                                                                                                                                                   \\
14                              & NingLab/ECInstruct      & Query Product Rank                & retrieval                & v7, v8           & 10000                             & Yes          &                                                                                                                                                                                                                                                                                                                                                                                                                                                                                   \\
15                              & NingLab/ECInstruct      & Sequential Recommendation         & multiple-choice          & v7, v8           & 10000                             & Yes          &                                                                                                                                                                                                                                                                                                                                                                                                                                                                                   \\
16                              & NingLab/ECInstruct      & Answerability Prediction          & multiple-choice          & v7, v8           & 10000                             & No           &                                                                                                                                                                                                                                                                                                                                                                                                                                                                                   \\
17                              & NingLab/ECInstruct      & Product Matching                  & multiple-choice          & v7, v8           & 4044                              & No           &                                                                                                                                                                                                                                                                                                                                                                                                                                                                                   \\
18                              & NingLab/ECInstruct      & Product Substitute Identification & multiple-choice          & v7, v8           & 10000                             & No           &                                                                                                                                                                                                                                                                                                                                                                                                                                                                                   \\
19                              & NingLab/ECInstruct      & Sentiment Analysis                & multiple-choice          & v7, v8           & 10000                             & No           &                                                                                                                                                                                                                                                                                                                                                                                                                                                                                   \\
20                              & Amazon-M2               & New Idea                                       & generation               & v7, v8           & 10000                             & Yes          & Explain product type given title, description, and product type                                                                                  \\
21                              & ESCI-data               & New Idea                                       & multiple-choice          & v7, v8           & 10000                             & Yes          & Select the user query that matches the product description                                                                                                                                                                                                                                                                                                                                                                                                                        \\
22                              & ESCI-data               & New Idea                                       & multiple-choice          & v7, v8           & 10000                             & Yes          & Select the user query that matches the product features                                                                                                                                                                                                                                                                                                                                                                                                                           \\
23                              & ESCI-data               & New Idea                                       & multiple-choice          & v7, v8           & 10000                             & Yes          & Select the user query that matches the product title                                                                                                                                                                                                                                                                                                                                                                                                                              \\
24                              & ESCI-data               & New Idea                                       & multiple-choice          & v7, v8           & 10000                             & Yes          & Select the title for the product description                                                                                                                                                                                                                                                                                                                                                                                                                                      \\
25                              & ESCI-data               & New Idea                                       & multiple-choice          & v7, v8           & 10000                             & Yes          & Select the title for the product features                                                                                                                                                                                                                                                                                                                                                                                                                                         \\
26                              & ESCI-data               & New Idea                                       & multiple-choice          & v7, v8           & 10000                             & Yes          & Select the product title for the user query                                                                                                                                                                                                                                                                                                                                                                                                                                       \\
27                              & ESCI-data               & New Idea                                       & retrieval                & v8               & 10000                             & No           & Pick 3 bullet points to match product                                                                                                                                                                                                                                                                                                                                                                                                                                             \\
28                              & NingLab/ECInstruct      & New Idea                                       & ranking                  & v8               & 5000                              & No           & Rank product reviews - positive to negative                                                                                                                                                                                                                                                                                                                                                                                                                                       \\
29                              & ESCI-data               & New Idea                                       & multiple-choice          & v8               & 5435                              & No           & Pick product to match query                                                                                                                                                                                                                                                                                                                                                                                                                                                       \\
30                              & ESCI-data               & New Idea                                       & ranking                  & v8               & 5000                              & No           & Task12 backwards. Given product, rank queries                                                                                                                                                                                                                                                                                                                                                                                                                                     \\
31                              & KDD Cup 2023            & New Idea                                       & retrieval                & v8               & 10000                             & No           & Given purchase pick previous clicks (similar to task 14)                                                                                                                                                                                                                                                                                                                                                                                                                          \\
32                              & ESCI-data               & New Idea                                       & multiple-choice          & v8               & 10000                             & No           & Given title pick brand                                                                                                                                                                                                                                                                                                                                                                                                                                                            \\
33                              & ESCI-data               & New Idea                                       & multiple-choice          & v9b              & 10000                             & No           & Given query product pair, what is relationship? E S C I                                                                                                                                                                                                                                                                                                                                                                                                                           \\
34                              & ESCI-data               & New Idea                                       & ranking                  & v9b              & 10000                             & No           & Given list of query product pairs, rank which are most related to least related                                                                                                                                                                                                                                                                                                                                                                                                   \\
35                              & ESCI-data               & New Idea                                       & retrieval                & v9b              & 10000                             & No           & Given query, select products which are exact match not substitute, complement, or irrelevent                                                                                                                                                                                                                                                                                                                                                                                      \\
36                              & Amazon Reviews 2023     & New Idea                                       & ranking                  & v9b              & 10000                             & No           & Given a product title and multiple reviews, rank the reviews based on the helpfulness                                                                                                                                                                                                                                                                                                                                                                                              \\
37                              & Alpaca Cleaned          & No Changes                          & generation               & v7, v8           & 51760                             & No           &                                                                                                                                                                                                                                                                                                                                                                                                                                                                                   \\
38                              & MMLU                    & No Changes                          & multiple-choice          & v7, v7b, v8      & 115700                            & No           &                                                                                                                                                                                                                                                                                                                                                                                                                                                                                   \\ \hline
\multicolumn{1}{l}{}            & Total                   & Total                                          & Total                    & Total            & 502435                            &              &                                                                                                                                                                                                                                                                                                                                                                                                                                                                                  
\end{tabular}
\label{tab:training_dataset}
\end{table}

\end{document}